\documentclass{article}
\usepackage{ijcai21}

\usepackage{times}
\usepackage{soul}
\usepackage{url}
\usepackage[hidelinks]{hyperref}
\usepackage[utf8]{inputenc}
\usepackage[small]{caption}
\usepackage{graphicx}
\usepackage{amsmath}
\usepackage{amsthm}
\usepackage{booktabs}
\usepackage{algorithm}
\usepackage{algorithmic}
\usepackage{graphicx}
\usepackage{amsmath}
\usepackage{algorithm,algorithmic}
\usepackage{amsfonts}
\usepackage{float}
\title{Box-Adapt: Domain-Adaptive Medical Image Segmentation using Bounding Box Supervision}

 \author{
 Yanwu XU$^1$
 \and
 Mingming Gong$^2$\and
 Shaoan Xie$^3$\and
 Kayhan Batmanghelich$^{1}$
 \affiliations
 $^1$Department of Biomedical Informatics, University of Pittsburgh, Pittsburgh , Pennsylvania , USA \\
 $^2$School of Mathematics and Statistics,
Melbourne Centre for Data Science,
The University of Melbourne, Australia\\
 $^3$Department of Philosophy,
Carnegie Mellon University\\
 \emails
 \{yanwuxu\}@pitt.edu
 }
\begin{document}

\maketitle

\begin{abstract}

Deep learning has achieved remarkable success in medical image segmentation, but it usually requires a large number of images labeled with fine-grained segmentation masks, and the annotation of these masks can be very expensive and time-consuming. Therefore, recent methods try to use unsupervised domain adaptation (UDA) methods to borrow information from labeled data from other datasets (source domains) to a new dataset (target domain). However, due to the absence of labels in the target domain, the performance of UDA methods is much worse than that of the fully supervised method. In this paper, we propose a weakly supervised domain adaptation setting, in which we can partially label new datasets with bounding boxes, which are easier and cheaper to obtain than segmentation masks. Accordingly, we propose a new weakly-supervised domain adaptation method called Box-Adapt, which fully explores the fine-grained segmentation mask in the source domain and the weak bounding box in the target domain. Our Box-Adapt is a two-stage method that first performs joint training on the source and target domains, and then conducts self-training with the pseudo-labels of the target domain. 
We demonstrate the effectiveness of our method in the liver segmentation task.
\end{abstract}

\section{Introduction}

In recent years, deep learning techniques  have become the de facto standard for various medical image analysis tasks \cite{medical_suvey}. Among these tasks, medical image segmentation, which aims to predict accurate masks for certain organs, is a key step to automatic disease analysis and surgeries. The deep CNN-based methods \cite{unet,vnet} have achieved significant advancements in brain tumor segmentation, liver segmentation \cite{brats,chaos}, etc. However, labeling medical images with fine-grained segmentation masks can be very expensive and time-consuming.  

\begin{figure}
\centering\includegraphics[width=0.5\textwidth]{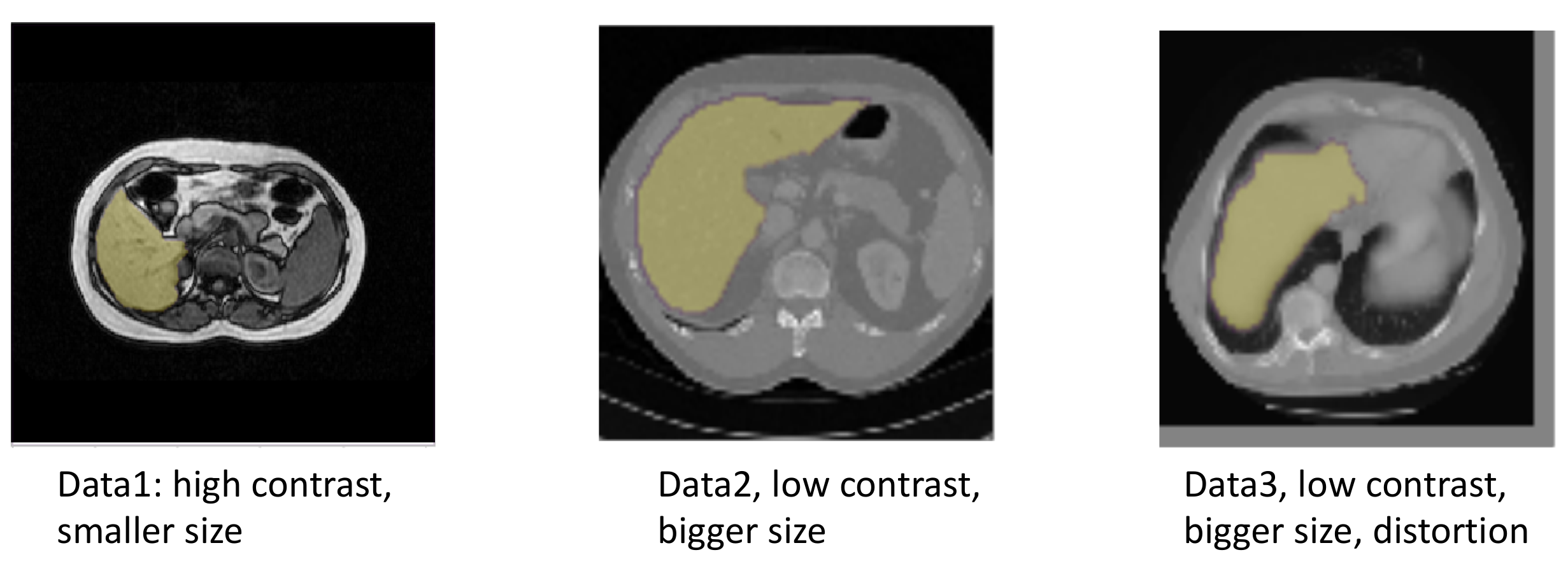}
\caption{Three random samples from three different dataset, the data distribution varies and the label could shift from one dataset to another dataset.} \label{data_shift}
\end{figure}

Given a new unlabeled dataset, one promising way is to borrow information from previous labeled datasets (source domains), for example, training a model using source domain data and apply it on the new test dataset (target domain). However, since the source domain and the target domain data do not necessarily have the same distribution, the deep model trained on the source domain may not generalize well to the target domain. For example, as shown in Fig.~\ref{data_shift}, different liver scanning from different datasets can exhibit drastic difference in intensity distribution and liver size. To mitigate the domain gap and improve the generalization ability, unsupervised domain adaptation (UDA) methods learn domain-adaptive classifiers by training on labeled source domain and unlabeled target domain data. These methods either learn domain-invariant representations \cite{da_adv,seg_adv,contrastive_uda} or translate source domain images to the style of the target domain images \cite{cyc_da,flow_da,multi_da,one_side_da}. However, because the target domain is unlabeled, the unsupervised learning methods cannot effectively align the conditional distribution of segmentation labels given images. As a consequence, the performance of UDA methods can be much worse than the supervised classifier and sometimes is even worse than the classifier trained on the source domain.

To alleviate the drawbacks of UDA methods and enable reliable adaptation, we propose to utilize some kind of weak supervision from the target domain. In the context of medical image segmentation, we can let the experts or even non-experts annotate bounding boxes that cover the objects in the target domain. In this way, we only need to annotate a few randomly selected slices of each randomly selected subject. Then, we make use of the bounding boxes in the target to develop a weakly-supervised domain adaptation (WDA) method called Box-Adapt, which also explores the weak bounding box annotations. Our method consists of two stages. In the first stage, we train the source and target segmentation networks jointly, using the source domain segmentation masks and the target domain bounding boxes as supervision. In the source domain, we use the standard cross-entropy loss, while in the target domain, we employ a non-negative Positive-Unlabeled (PU) loss \cite{non_pu,3D_bbox}, which provides unbiased estimation of target domain segmentation error from bounding box supervision. The joint training helps align the conditional distribution of label given images, and thus help adjust source domain segmentation network to the target domain. In the second stage, we use the network learned from the first stage to generate pseudo-labels on the target domain unlabeled slices and refine the network using such pseudo labels, which lead to further adaptation of the segmentation network to the target domain. 

Our contributions can be summarized as follows:

\begin{description}
  \item[$\bullet$] To the best of our knowledge, we are the first to propose the WDA setting with bounding box weak supervision for medical image segmentation.
  \item[$\bullet$] We propose a WDA method called Box-Adapt for learning accurate segmentation model with fully labeled source domain  data and weakly annotated target domain.
  \item[$\bullet$] We demonstrate the effectiveness of our method on a liver segmentation dataset.
\end{description}

\section{Box-Adapt for Medical Image Segmentation}\label{method}

\begin{figure*}
\centering
\includegraphics[width=0.8\textwidth]{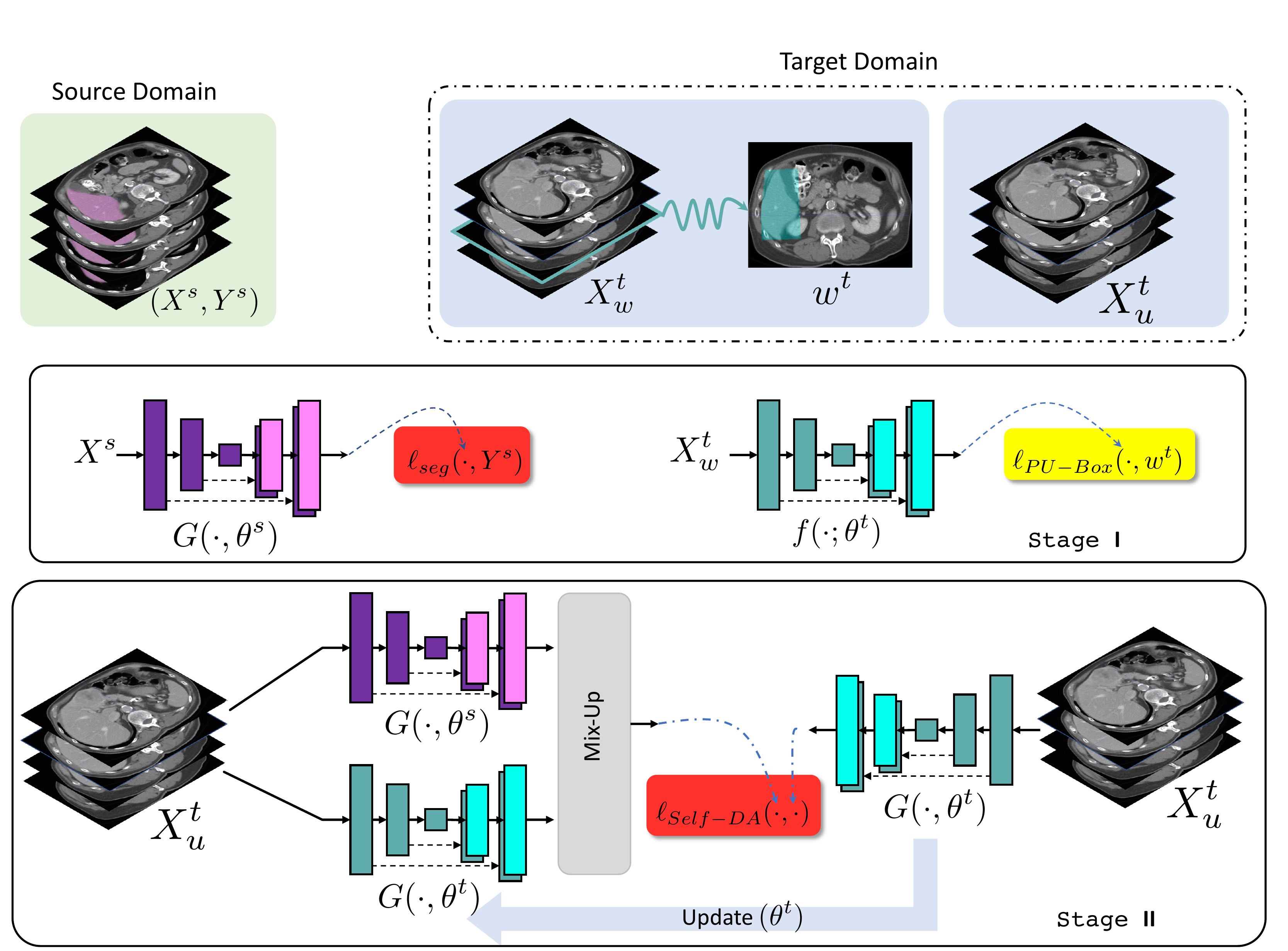}
\caption{overall model of our proposed Box-Adapt method.} \label{model}
\end{figure*}
In this section, we first introduce our weakly-supervised domain adaptation setup. Then, we present the details of our Box-Adapt method.

\subsection{Problem Setting}
In the proposed weakly-supervised domain adaptation scenario, we have access to a labeled source domain $\mathcal{S} = \{x^s_i,y^s_i\}^n_{i=1}$, where $y^s_i$ denotes a ground-truth segmentation mask labeled by medical experts. In the target domain, we have unlabeled data $\mathcal{T}_u=\{x^t_i\}^l_{i=1}$ weakly-labeled data $\mathcal{T}_l=\{x^t_i,w^t_i\}^m_{i=1}$, where $w^t_i$ is the bounding box annotation. 
Our goal is to learn from $\mathcal{S}$, $\mathcal{T}_u$, and $\mathcal{T}_l$ a network $G_t$ that produces accurate pixel-wise labels in the target domain.

\subsection{Box-Adapt}

Our Box-Adapt method aims to learn the target domain segmentation network $G_t$ by exploiting the information in $\mathcal{S}$, $\mathcal{T}_u$, and $\mathcal{T}_l$ in a unified framework. To aid the training of $G_t$, we add a segmentation network $G_s$, which is trained by labeled data $\mathcal{S}$ from the source domain. Our method consists of two stages. In the first stage, we train $G_s$ and $G_t$ jointly with $\mathcal{S}$ and $\mathcal{T}_l$, where $G_s$ is trained on $\mathcal{S}$ and $G_t$ is trained on $\mathcal{T}_l$. We let $G_s$ and $G_t$ share the convolutional weights, such that the feature extraction layer in $G_t$ can also be supervised source domain segmentation masks, which enables knowledge transfer from the source domain. In the second stage, we use $G_s$ and $G_t$ to generate pseudo labels on $\mathcal{T}_u$ and $\mathcal{T}_l$, and then mix up the pseudo labels from $G_s$ and $G_t$ into a new form of pseudo label to refine $G_t$ for the target domain. The overall architecture of our approach is shown in Fig. \ref{model}. In the following, we provide technical details of the two stages.

\subsubsection{Stage I} 
In this stage, we train $G_s$ and $G_t$ jointly with $\mathcal{S}$ and $\mathcal{T}_l$. We let $G_s$ and $G_t$ share the weights in convolutional layers, such that the feature extraction layer in $G_t$ can be learned by using source domain segmentation mask supervision. Because source domain has pixel-level segmentation mask annotations, we train $G_s$ using the pixel-wise cross-entropy loss:
\begin{equation}
    \ell_{seg} = -\frac{1}{HL}\sum^{HL}_{k=1}y_{ik}^s \log(G_s(x_i^s)_{k}),
\end{equation}
where $G_s(x_i^s)_{k}$ and $y_{ik}^s$ represent the network output and the label for one of the $k$-th pixel in $i$-th training pair $(x_i^s,y_i^s)$ and $HL$ is the number of pixels in the image. 
In the target domain, since we only have bounding box annotations, a naive way is to set the pixels inside the bounding boxes as foreground and the rest pixels as background and use cross-entropy loss to train $G_t$. However, because the bounding box usually contains a small proportion of background pixels, training with it will lead to a biased classifier in the target domain. Here we consider bounding box supervised learning as a positive-unlabeled (PU) learning problem, where the pixels outside the boxes are considered as positive examples and the pixels within the boxes are considered as unlabeled examples. We thus employ a PU learning loss \cite{kiryo2017positive} for training $G_t$:
\begin{align}\label{pu}
    \ell_{Pu-Box} &= \frac{1}{HL} (\pi_p\sum^{HL}_{k=1}(1-w_{ik}^t) \log(1-G_t(x_i^t)_{k}) \nonumber\\
    & + \max\{0 ,\sum^{HL}_{k=1}w_{ik}^t \log(G_t(x_i^t)_{k}) \\
&-\pi_p\sum^{HL}_{k=1}w_{ik}^t \log(1-G_t(x_i^t)_{k})  \}),
\end{align}
where $G_t(x_i^t)_{k}$ and $y_{ik}^t$ represent the network output and the label for one of the $k$-th pixel in $i$-th training pair $(x_i^t,w_i^t)$ and $HL$ is the number of pixels in the image. The hyperparameter $1-\pi_p$ denotes the proportion of the positive segmentation map with respect to the background over all the dataset. Previously works in PU learning assume that $\pi_p$ is a known prior, but in realistic application this could be unpredictable, especially for the noisy medical data. Thanks to the source domain data, we propose to estimate the $\pi_p$ dynamically for each $x_i^t$ by using the source domain segmentation networks:
\begin{equation}\label{pi}
    \hat{\pi}_p^i = \frac{1 - \sum^{HL}_{k=1}G_s(x_i^t)_{k}}{HL}.
\end{equation}
Though the prediction might be coarse at the beginning, the prediction will be successively refined during the training. The overall loss for stage one is the combination of source and target domain losses, i.e., $\ell_{Stage I} = \ell_{seg} + \ell_{Pu-Box}$.

\subsubsection{Stage II} Thanks to the joint training strategy \cite{self_effective,simple_self}, we can obtain a learned $G_t$ that works well in the target domain in the first stage. To further improve the generalization of $G_t$, we propose a second-stage refinement of $G_t$ by self-training, which was originally developed for semi-supervised learning and then employed for unsupervised domain adaptation. Different from previous approaches, we use $G_s$ as a reference network to construct pseudo labels together with $G_t$ by mixing up the predictions, leading to more reliable pseudo labels. In addition, we make use of available bounding boxes in the target domain to refine the pseudo labels, which further reduces the negative effects caused by noisy pseudo labels.

In specific, for the unlabeled target domain data $\mathcal{T}_u$, we use $G_s$ and $G_t$ to generate pseudo labels and then mix them up to obtain the final pseudo labels which can be used to update $G_t$ as follows
\begin{align}\label{self_da}
    \ell_{Self-DA} &= \frac{1}{HL}\sum^{HL}_{k=1}((1-\alpha)G_s(x_i^t)_k \nonumber\\
    &+\alpha G_t(x_i^t)_k)\log(G_t(x_i^t)_k),
\end{align}
where $\alpha$ is a parameter weighting contributions from $G_s$ and $G_t$. For the weakly-labeled target domain data $\mathcal{T}_l$, we employ the same procedure, except that we utilize the bounding box annotations to refine the pseudo labels by taking the intersection between the pseudo segmentation mask and the bounding box in an image:
\begin{align}\label{self-box}
    \ell_{Self-Box} &= \nonumber\\ &-\frac{1}{HL}\sum^{HL}_{k=1}([G_s(x_i^t)_k=1]_+[w_{ik}^t=1]_+\log(G_t(x_i^t)_k) \nonumber\\ &+ [w_{ik}^t=0]\log(G_t(x_i^t)_k)).
\end{align}

In equation~\ref{self-box}, we use the symbol function $[\cdot]_+$ to constrain the prediction of $G_s$, where we only use the prediction of $G_s$ inside the area of bounding box. Overall, in the second training stage, the training objective is the sum of $\ell_{Self-DA}$ $\ell_{Self-Box}$.


\subsubsection{Discussion} In the computer vision community, \cite{weak_da} proposes a WDA approach for semantic segmentation. Our method is different from \cite{weak_da} in two aspects. First, \cite{weak_da} utilizes the image-level category labels, but we use the bounding box annotations as our weak supervision, which is more practical for medical image segmentation. Second, their method is built upon adversarial learning, but we construct our WDA method via a two-step strategy. \cite{3D_bbox} proposes a PU learning method to learn segmentation masks from bounding box annotations without the aid of a source domain. However, as shown in our experiments, solely relying on bounding box supervision is insufficient to obtain accurate segmentation masks. With the source domain's help, we can estimate the prior $\pi_p$ in PU loss and transfer knowledge to achieve more accurate segmentation.
\section{Experiments}\label{experiments}
\subsection{Training Settings}
We apply Resnet101 \cite{resnet} as our backbone for 2D medical image segmentation. To save the training memory and boost the training speed of models, we share most of the parameters between $G_s$ and $G_t$ until the fourth res-block of Resnet101. For the last layer of $G_s$ and $G_t$, we add another down-sampling res-block for them respectively. In the training stage one, We train the model jointly with the segmentation loss $\ell_{seg}$ and bounding box loss $\ell_{pu-box}$. The weight between $\ell_{seg}$ and $\ell_{pu-box}$ are both $1.0$. In the second training stage, we load the $G_s$ and $G_t$ with the initial parameters which are learned from stage I. The training parameters of $G_s$ and $G_t$ are not shared anymore. Also, as motioned in the Section~\ref{method}, the parameters of $G_s$ are fixed for providing unchanged pseudo masks of segmentation. Then, we fuse the pseudo masks generated from $G_s$ and $G_t$ for updating the $G_t$. $G_t$ is updated for each batch, thus the generated pseudo masks are improved with the updated $G_t$. To train our proposed method and the referred adversarial method  \cite{seg_adv}, we adopt ADAM \cite{adam} optimizer with learning rate $0.0002$ and $\beta = {0.5,0.999}$, which is one of the common settings used in medical segmentation tasks. During the training, we set the batch size to be 4 for all implementation and run each setting for 20000 iterations with a decay of learning rate by 0.9. All models are initialized with the weights pre-trained from the ImageNet dataset \cite{imagenet}.

\subsection{Dataset}
We focus on the liver segmentation task and use two benchmark dataset CHAOS  \cite{chaos}, SLIVER \cite{sliver} and the recent MedSeg  \cite{medseg}. The CHAOS dataset consists two modalities of CT and MRI. For the CT of CHAOS, it consists of data from three different scanning situations and has in total 40 different patients. For the MRI images, there are 120 different patient with both of the T2-weighted and T1-DUAL modalities. In the dataset of SLIVER and MedSeg, they all only include the CT scannings with 20 and 50 different patients. At the data prepossessing stage, we apply the uniform strategy to align all the 3D scanning from each dataset in the same direction and resize the resolution to be $256\times 256$. The density of all data are restricted to be in the range of $[0,1400]$ and then is normalized in the range of $[0,255]$ for training the models. We conduct two directions of domain adaptation, which are adapting from CHAOS to SLIVER and CHAOS to MedSeg. To simulate the realistic scenario of applications, we only use $25\%$ of CHAOS as the source data. For the target dataset SLIVER and MedSeg, we randomly sample $80\%$ of data for our WDA and the rest is for evaluation. To generate the coarse bounding box label in the SLIVER and MedSeg dataset, we extract only three center slices from each patients and annotate them with bound box mask to cover the liver. And in the real applications, annotate bounding box annotation for each slice only cost several seconds, which is considerably fast.

\subsection{Experimental Results}\label{experiment_results}
\begin{figure*}[h]
\centering
\begin{minipage}{.5\textwidth}
  \centering
  \includegraphics[width=0.9\linewidth]{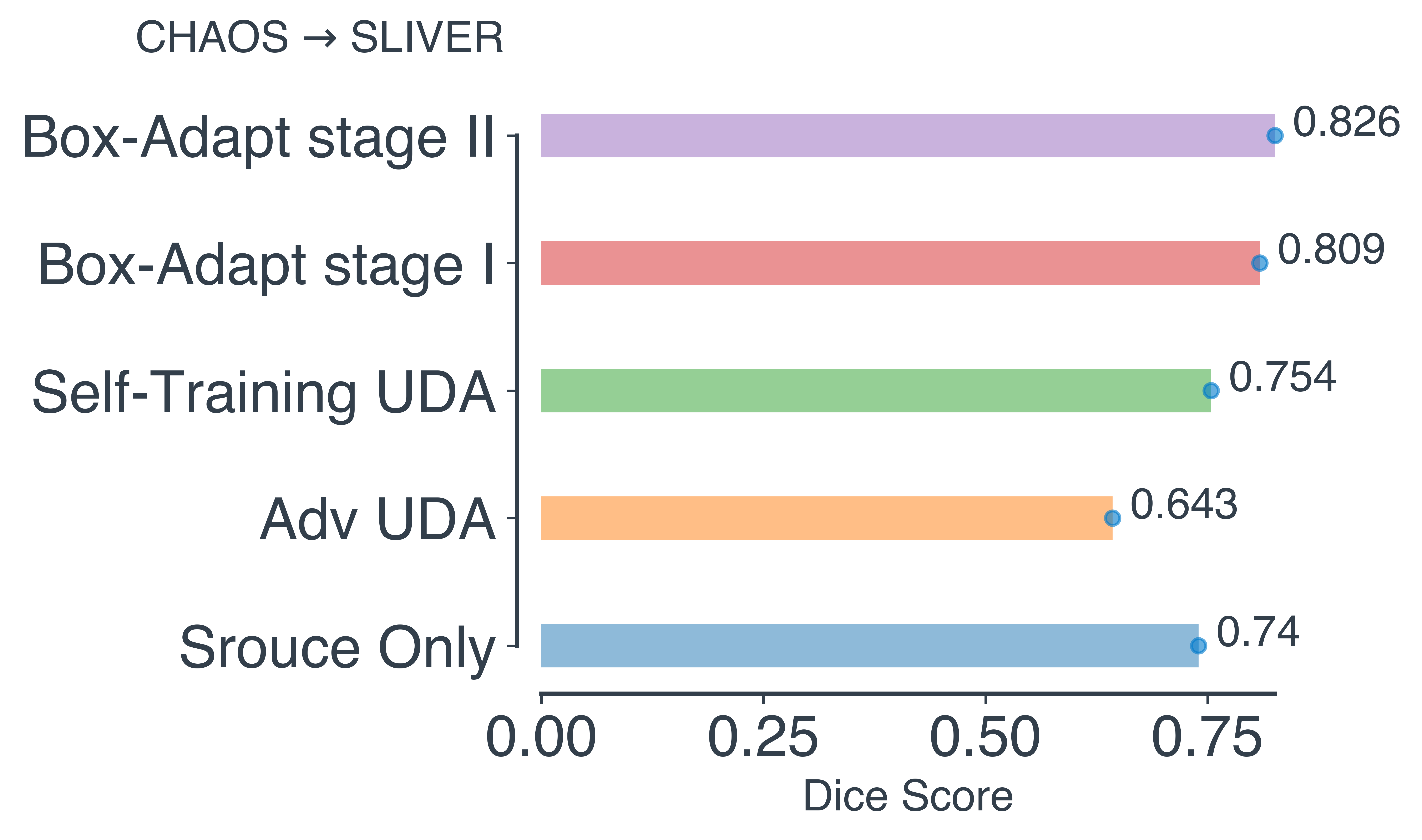}
  \caption{Dice score of CHAOS $\to$ SLIVER.}
  \label{fig:sliver}
\end{minipage}%
\begin{minipage}{.5\textwidth}
  \centering
  \includegraphics[width=0.9\linewidth]{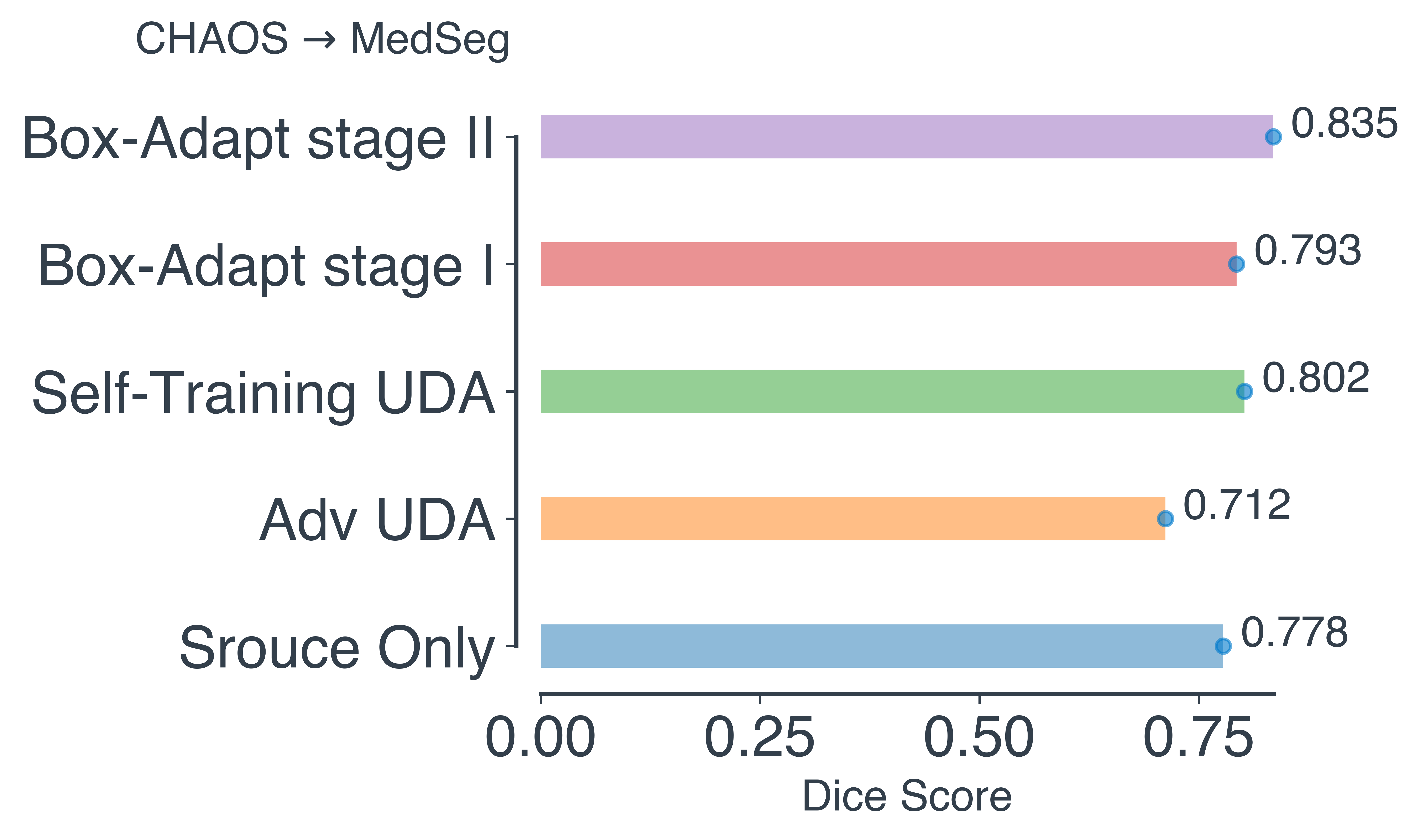}
  \caption{Dice score of CHAOS $\to$ MedSeg.}
  \label{fig:medseg}
\end{minipage}
\end{figure*}

We use the Dice Score for our our final evaluation on the extra testing data, to be notified, all of the testing data does not join the training process. For the comparison, we run five settings, the source only: model is only trained with annotated source data; Adv UDA represents the method of  \cite{seg_adv}; Box-Adapt: we run our model without any label information from target domain, which drops the loss of $\ell_{PU-Box}$ and $\ell_{Self-Box}$ in the training stage one and stage 2. In this case, the stage one for Box-Adapt is just source only model; Our Box-Adapt stage one: it represents the results of our proposed method in stage one; Our Box-Adapt DA stage two: the results is generated by our proposed method of stage two.

\begin{figure*}
\includegraphics[width=1\textwidth]{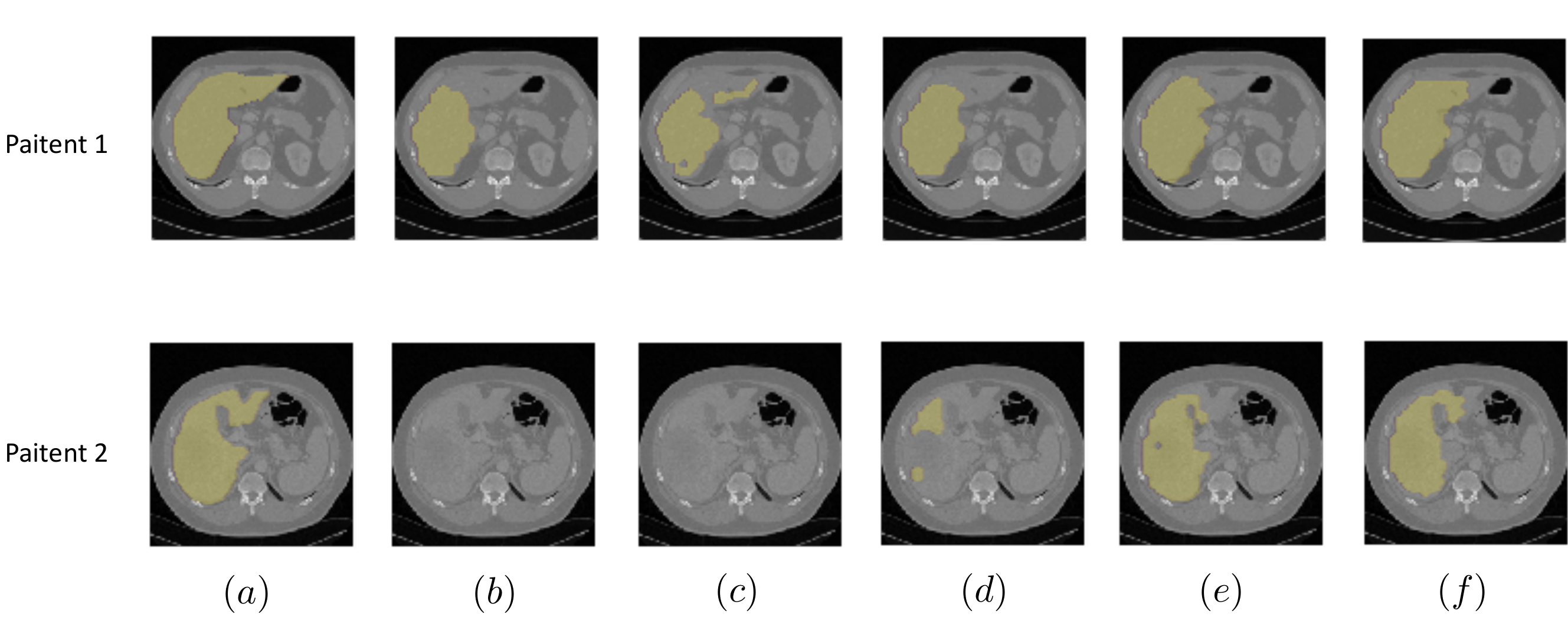}
\caption{(a) Ground truth segmentation, (b) Source only, (c) Adv UDA, (d) Self-Training UDA, (e) Box-Adapt (ours) with less bounding box annotation (3 slices each patient), (f) Box-Adapt (ours) with more bounding box annotation (10 slices each patient). } \label{visual_result}
\end{figure*}

The results for the adaptation scenario of CHAOS $\to$ SLIVER and CHAOS $\to$ MedSeg are shown in Fig.~\ref{fig:sliver} and Fig.~\ref{fig:medseg}. We observed that the adversarial method could hurt the performance in our case which drops the prediction performance of the segmentation model. For the Box-Adapt without bounding box learning, it can benefit from the self-training process slightly by improving $1.5\%$ accuracy over dice score. Once, we utilize the course information from several annotations from bounding box, the performance improves by a big gap compared to the source only model and the Box-Adapt model. Also, in our training stage two, our proposed method also achieves the improvements of $3\%$ compared to the stage one. Visually, we show the example of segmentation output from different models in Fig.~\ref{visual_result}, our proposed method improves the quality of the segmentation masks from stage one to stage two also with the help of bounding box annotations, however, the source only model and the adversarial UDA fails to find the location of liver in some cases. For more results, it can be refereed in the supplementary section. The results have shown the effectiveness of our model compared to the traditional UDA method with effortless weak information in the target domain.

\subsection{Ablation Study of the Numbers of Weak Labels}
\begin{table}
	\centering
	\caption{CHAOS $\to$ SLIVER.}\label{ablation}
	\begin{tabular}{|l|c|c|c|c|c|c|c|c}
		\hline
		& \multicolumn{2}{c|}{0 slice}& \multicolumn{2}{c|}{1 slice} & \multicolumn{2}{c|}{3 slice} & \multicolumn{2}{c|}{10 slice} \\ \hline
		Box-Adapt stage I & \multicolumn{2}{c|}{0.740} &\multicolumn{2}{c|}{0.761} & \multicolumn{2}{c|}{0.814} & \multicolumn{2}{c|}{0.861} \\
		Box-Adapt stage II & \multicolumn{2}{c|}{0.754} & \multicolumn{2}{c|}{0.782} & \multicolumn{2}{c|}{0.820} & \multicolumn{2}{c|}{0.870} \\
		\hline
	\end{tabular}
\end{table}
In this section, we study the effectiveness of the bounding box annotations. The results of Section~\ref{experiment_results} consider the setting of three slices of bounding box annotations for each patient on CHAOS $\to$ SLIVER. Here, we only set the number of annotations for each patient to be 0,1,3,10 for comparison due to the limited resource of computation. In Table~\ref{ablation}, 0 slice means source only model in the stage one and Box-Adapt in stage two. Basically, the performance of the model continues to improve as the number of annotation grows, and 3 slices bounding box annotations has shown to be the most effective one with much less annotation compared with 10 slices and improves the performance by a big gap compared with one slice.

\section{Conclusion}
In this paper, we propose a novel framework of the weakly supervised domain adaptation(WDA) for the medical image segmentation. In our model, we propose to utilize the partially labeled bounding box annotation in the new domain to enhance the domain adaptation. We construct our model by two-stage learning, the first stage jointly trains the model on the source domain and the target domain, in the second stage, we conduct self-training on the target box-annotated data and the unlabeled data. We compared our method with the the adversarial domain adaptation method and the method without bounding box annotations on liver segmentation tasks. Our proposed Box-Adapt framework shows a favorable results throughout.

\bibliographystyle{named}
\bibliography{ijcai21.bbl}

\begin{thebibliography}{}

\bibitem[\protect\citeauthoryear{Bakas \bgroup \em et al.\egroup
  }{2018}]{brats}
Spyridon Bakas, Mauricio Reyes, Andr{\'{a}}s Jakab, Stefan Bauer, Markus
  Rempfler, Alessandro Crimi, Russell~Takeshi Shinohara, Christoph Berger,
  Sung~Min Ha, Martin Rozycki, Marcel Prastawa, Esther Alberts, Jana
  Lipkov{\'{a}}, John~B. Freymann, Justin~S. Kirby, Michel Bilello, Hassan~M.
  Fathallah{-}Shaykh, Roland Wiest, Jan Kirschke, Benedikt Wiestler, Rivka~R.
  Colen, Aikaterini Kotrotsou, Pamela LaMontagne, Daniel~S. Marcus, Mikhail
  Milchenko, Arash Nazeri, Marc{-}Andr{\'{e}} Weber, Abhishek Mahajan, Ujjwal
  Baid, Dongjin Kwon, Manu Agarwal, Mahbubul Alam, Alberto Albiol, Antonio
  Albiol, Alex Varghese, Tran~Anh Tuan, Tal Arbel, Aaron Avery, Pranjal B.,
  Subhashis Banerjee, Thomas Batchelder, Kayhan~N. Batmanghelich, Enzo
  Battistella, Martin Bendszus, Eze Benson, Jos{\'{e}} Bernal, George Biros,
  Mariano Cabezas, Siddhartha Chandra, Yi{-}Ju Chang, and et~al.
\newblock Identifying the best machine learning algorithms for brain tumor
  segmentation, progression assessment, and overall survival prediction in the
  {BRATS} challenge.
\newblock {\em CoRR}, abs/1811.02629, 2018.

\bibitem[\protect\citeauthoryear{Benaim and Wolf}{2017}]{one_side_da}
Sagie Benaim and Lior Wolf.
\newblock One-sided unsupervised domain mapping.
\newblock In I.~Guyon, U.~V. Luxburg, S.~Bengio, H.~Wallach, R.~Fergus,
  S.~Vishwanathan, and R.~Garnett, editors, {\em Advances in Neural Information
  Processing Systems}, volume~30. Curran Associates, Inc., 2017.

\bibitem[\protect\citeauthoryear{Deng \bgroup \em et al.\egroup
  }{2009}]{imagenet}
J.~Deng, W.~Dong, R.~Socher, L.-J. Li, K.~Li, and L.~Fei-Fei.
\newblock {ImageNet: A Large-Scale Hierarchical Image Database}.
\newblock In {\em CVPR09}, 2009.

\bibitem[\protect\citeauthoryear{Ganin \bgroup \em et al.\egroup
  }{2016}]{da_adv}
Yaroslav Ganin, Evgeniya Ustinova, Hana Ajakan, Pascal Germain, Hugo
  Larochelle, Fran{\c{c}}ois Laviolette, Mario March, and Victor Lempitsky.
\newblock Domain-adversarial training of neural networks.
\newblock {\em Journal of Machine Learning Research}, 17(59):1--35, 2016.

\bibitem[\protect\citeauthoryear{Gong \bgroup \em et al.\egroup
  }{2019}]{flow_da}
Rui Gong, Wen Li, Yuhua Chen, and Luc~Van Gool.
\newblock Dlow: Domain flow for adaptation and generalization.
\newblock In {\em Proceedings of the IEEE/CVF Conference on Computer Vision and
  Pattern Recognition (CVPR)}, June 2019.

\bibitem[\protect\citeauthoryear{{He} \bgroup \em et al.\egroup
  }{2016}]{resnet}
K.~{He}, X.~{Zhang}, S.~{Ren}, and J.~{Sun}.
\newblock Deep residual learning for image recognition.
\newblock In {\em 2016 IEEE Conference on Computer Vision and Pattern
  Recognition (CVPR)}, pages 770--778, 2016.

\bibitem[\protect\citeauthoryear{{Heimann} \bgroup \em et al.\egroup
  }{2009}]{sliver}
T.~{Heimann}, B.~{van Ginneken}, M.~A. {Styner}, Y.~{Arzhaeva}, V.~{Aurich},
  C.~{Bauer}, A.~{Beck}, C.~{Becker}, R.~{Beichel}, G.~{Bekes}, F.~{Bello},
  G.~{Binnig}, H.~{Bischof}, A.~{Bornik}, P.~M.~M. {Cashman}, Y.~{Chi},
  A.~{Cordova}, B.~M. {Dawant}, M.~{Fidrich}, J.~D. {Furst}, D.~{Furukawa},
  L.~{Grenacher}, J.~{Hornegger}, D.~{KainmÜller}, R.~I. {Kitney},
  H.~{Kobatake}, H.~{Lamecker}, T.~{Lange}, J.~{Lee}, B.~{Lennon}, R.~{Li},
  S.~{Li}, H.~{Meinzer}, G.~{Nemeth}, D.~S. {Raicu}, A.~{Rau}, E.~M. {van
  Rikxoort}, M.~{Rousson}, L.~{Rusko}, K.~A. {Saddi}, G.~{Schmidt},
  D.~{Seghers}, A.~{Shimizu}, P.~{Slagmolen}, E.~{Sorantin}, G.~{Soza},
  R.~{Susomboon}, J.~M. {Waite}, A.~{Wimmer}, and I.~{Wolf}.
\newblock Comparison and evaluation of methods for liver segmentation from ct
  datasets.
\newblock {\em IEEE Transactions on Medical Imaging}, 28(8):1251--1265, 2009.

\bibitem[\protect\citeauthoryear{Hoffman \bgroup \em et al.\egroup
  }{2018}]{cyc_da}
Judy Hoffman, Eric Tzeng, Taesung Park, Jun-Yan Zhu, Phillip Isola, Kate
  Saenko, Alexei Efros, and Trevor Darrell.
\newblock {C}y{CADA}: Cycle-consistent adversarial domain adaptation.
\newblock In Jennifer Dy and Andreas Krause, editors, {\em Proceedings of the
  35th International Conference on Machine Learning}, volume~80 of {\em
  Proceedings of Machine Learning Research}, pages 1989--1998,
  Stockholmsmässan, Stockholm Sweden, 10--15 Jul 2018. PMLR.

\bibitem[\protect\citeauthoryear{Hosseini{-}Asl \bgroup \em et al.\egroup
  }{2018}]{multi_da}
Ehsan Hosseini{-}Asl, Yingbo Zhou, Caiming Xiong, and Richard Socher.
\newblock A multi-discriminator cyclegan for unsupervised non-parallel speech
  domain adaptation.
\newblock {\em CoRR}, abs/1804.00522, 2018.

\bibitem[\protect\citeauthoryear{Jenssen}{2020}]{medseg}
Tomas Jenssen, Håvard Bjørke;~Sakinis.
\newblock Medseg; jenssen, håvard bjørke; sakinis, tomas.
\newblock 2020.

\bibitem[\protect\citeauthoryear{{Kang} \bgroup \em et al.\egroup
  }{2019}]{contrastive_uda}
G.~{Kang}, L.~{Jiang}, Y.~{Yang}, and A.~G. {Hauptmann}.
\newblock Contrastive adaptation network for unsupervised domain adaptation.
\newblock In {\em 2019 IEEE/CVF Conference on Computer Vision and Pattern
  Recognition (CVPR)}, pages 4888--4897, 2019.

\bibitem[\protect\citeauthoryear{Kavur \bgroup \em et al.\egroup
  }{2021}]{chaos}
A.~Emre Kavur, N.~Sinem Gezer, Mustafa Barış, Sinem Aslan, Pierre-Henri
  Conze, Vladimir Groza, Duc~Duy Pham, Soumick Chatterjee, Philipp Ernst,
  Savaş Özkan, Bora Baydar, Dmitry Lachinov, Shuo Han, Josef Pauli, Fabian
  Isensee, Matthias Perkonigg, Rachana Sathish, Ronnie Rajan, Debdoot Sheet,
  Gurbandurdy Dovletov, Oliver Speck, Andreas Nürnberger, Klaus~H. Maier-Hein,
  Gözde {Bozdağı Akar}, Gözde Ünal, Oğuz Dicle, and M.~Alper Selver.
\newblock {CHAOS Challenge - combined (CT-MR) healthy abdominal organ
  segmentation}.
\newblock {\em Medical Image Analysis}, 69:101950, April 2021.

\bibitem[\protect\citeauthoryear{Kingma and Ba}{2014}]{adam}
Diederik~P. Kingma and Jimmy Ba.
\newblock Adam: {A} method for stochastic optimization.
\newblock {\em CoRR}, abs/1412.6980, 2014.

\bibitem[\protect\citeauthoryear{Kiryo \bgroup \em et al.\egroup
  }{2017a}]{non_pu}
Ryuichi Kiryo, Gang Niu, Marthinus~C du~Plessis, and Masashi Sugiyama.
\newblock Positive-unlabeled learning with non-negative risk estimator.
\newblock In I.~Guyon, U.~V. Luxburg, S.~Bengio, H.~Wallach, R.~Fergus,
  S.~Vishwanathan, and R.~Garnett, editors, {\em Advances in Neural Information
  Processing Systems}, volume~30. Curran Associates, Inc., 2017.

\bibitem[\protect\citeauthoryear{Kiryo \bgroup \em et al.\egroup
  }{2017b}]{kiryo2017positive}
Ryuichi Kiryo, Gang Niu, Marthinus~C du~Plessis, and Masashi Sugiyama.
\newblock Positive-unlabeled learning with non-negative risk estimator.
\newblock In {\em Proceedings of the 31st International Conference on Neural
  Information Processing Systems}, pages 1674--1684, 2017.

\bibitem[\protect\citeauthoryear{Lee}{2013}]{simple_self}
Dong-Hyun Lee.
\newblock Pseudo-label : The simple and efficient semi-supervised learning
  method for deep neural networks.
\newblock {\em ICML 2013 Workshop : Challenges in Representation Learning
  (WREPL)}, 07 2013.

\bibitem[\protect\citeauthoryear{Litjens \bgroup \em et al.\egroup
  }{2017}]{medical_suvey}
Geert Litjens, Thijs Kooi, Babak~Ehteshami Bejnordi, Arnaud Arindra~Adiyoso
  Setio, Francesco Ciompi, Mohsen Ghafoorian, Jeroen~A.W.M. {van der Laak},
  Bram {van Ginneken}, and Clara~I. Sánchez.
\newblock A survey on deep learning in medical image analysis.
\newblock {\em Medical Image Analysis}, 42:60--88, 2017.

\bibitem[\protect\citeauthoryear{McClosky \bgroup \em et al.\egroup
  }{2006}]{self_effective}
David McClosky, Eugene Charniak, and Mark Johnson.
\newblock Effective self-training for parsing.
\newblock In {\em Proceedings of the Human Language Technology Conference of
  the {NAACL}, Main Conference}, pages 152--159, New York City, USA, June 2006.
  Association for Computational Linguistics.

\bibitem[\protect\citeauthoryear{{Milletari} \bgroup \em et al.\egroup
  }{2016}]{vnet}
F.~{Milletari}, N.~{Navab}, and S.~{Ahmadi}.
\newblock V-net: Fully convolutional neural networks for volumetric medical
  image segmentation.
\newblock In {\em 2016 Fourth International Conference on 3D Vision (3DV)},
  pages 565--571, 2016.

\bibitem[\protect\citeauthoryear{Paul \bgroup \em et al.\egroup
  }{2020}]{weak_da}
Sujoy Paul, Yi-Hsuan Tsai, Samuel Schulter, Amit~K. Roy-Chowdhury, and Manmohan
  Chandraker.
\newblock Domain adaptive semantic segmentation using weak labels.
\newblock In {\em European Conference on Computer Vision (ECCV)}, 2020.

\bibitem[\protect\citeauthoryear{Ronneberger \bgroup \em et al.\egroup
  }{2015}]{unet}
Olaf Ronneberger, Philipp Fischer, and Thomas Brox.
\newblock U-net: Convolutional networks for biomedical image segmentation.
\newblock In Nassir Navab, Joachim Hornegger, William~M. Wells, and
  Alejandro~F. Frangi, editors, {\em Medical Image Computing and
  Computer-Assisted Intervention -- MICCAI 2015}, pages 234--241, Cham, 2015.
  Springer International Publishing.

\bibitem[\protect\citeauthoryear{Tsai \bgroup \em et al.\egroup
  }{2018}]{seg_adv}
Y.-H. Tsai, W.-C. Hung, S.~Schulter, K.~Sohn, M.-H. Yang, and M.~Chandraker.
\newblock Learning to adapt structured output space for semantic segmentation.
\newblock In {\em IEEE Conference on Computer Vision and Pattern Recognition
  (CVPR)}, 2018.

\bibitem[\protect\citeauthoryear{Xu \bgroup \em et al.\egroup }{2020}]{3D_bbox}
Yanwu Xu, Mingming Gong, Junxiang Chen, Ziye Chen, and Kayhan Batmanghelich.
\newblock 3d-boxsup: Positive-unlabeled learning of brain tumor segmentation
  networks from 3d bounding boxes.
\newblock {\em Frontiers in Neuroscience}, 14:350, 2020.

\end{thebibliography}

\end{document}